\documentclass[conference]{IEEEtran}
\IEEEoverridecommandlockouts
\usepackage{cite}
\usepackage{amsmath,amssymb,amsfonts}
\usepackage{algorithmic}
\usepackage{graphicx}
\usepackage{textcomp}
\usepackage{xcolor}
\usepackage{caption}
\usepackage{subcaption}
\def\BibTeX{{\rm B\kern-.05em{\sc i\kern-.025em b}\kern-.08em
    T\kern-.1667em\lower.7ex\hbox{E}\kern-.125emX}}
\usepackage{flushend}

\begin{document}

\title{Segmentation of Planning Target Volume in CT Series for Total Marrow Irradiation Using U-Net\\
\thanks{This work is supported by Italian Ministry of Health (grant number: GR-2019-12370739, project: AuToMI).}
}

\author{
\IEEEauthorblockN{Ricardo Coimbra Brioso\IEEEauthorrefmark{2}, Damiano Dei\IEEEauthorrefmark{3}\IEEEauthorrefmark{4}, Ciro Franzese\IEEEauthorrefmark{3}\IEEEauthorrefmark{4}, Nicola Lambri\IEEEauthorrefmark{3}\IEEEauthorrefmark{4},\\ Daniele Loiacono\IEEEauthorrefmark{2},
Pietro Mancosu\IEEEauthorrefmark{4}, and Marta Scorsetti\IEEEauthorrefmark{3}\IEEEauthorrefmark{4}}
\smallskip
\IEEEauthorblockA{\IEEEauthorrefmark{1}Dipartimento di Elettronica, Informazione e Bioingegneria, Politecnico di Milano, Milan, Italy}
\IEEEauthorblockA{\IEEEauthorrefmark{3}Department of Biomedical Sciences, Humanitas University, Pieve Emanuele, Milan, Italy}
\IEEEauthorblockA{\IEEEauthorrefmark{4}Radiotherapy and Radiosurgery Department, IRCCS Humanitas Research Hospital, Rozzano, Milan, Italy}
}


\maketitle

\begin{abstract}
    Radiotherapy (RT) is a key component in the treatment of various cancers, including Acute Lymphocytic Leukemia (ALL) and Acute Myelogenous Leukemia (AML). Precise delineation of organs at risk (OARs) and target areas is essential for effective treatment planning. Intensity Modulated Radiotherapy (IMRT) techniques, such as Total Marrow Irradiation (TMI) and Total Marrow and Lymph node Irradiation (TMLI), provide more precise radiation delivery compared to Total Body Irradiation (TBI). However, these techniques require time-consuming manual segmentation of structures in Computerized Tomography (CT) scans by the Radiation Oncologist (RO). In this paper, we present a deep learning-based auto-contouring method for segmenting Planning Target Volume (PTV) for TMLI treatment using the U-Net architecture. We trained and compared two segmentation models with two different loss functions on a dataset of 100 patients treated with TMLI at the Humanitas Research Hospital between 2011 and 2021. Despite challenges in lymph node areas, the best model achieved an average Dice score of 0.816 for PTV segmentation. Our findings are a preliminary but significant step towards developing a segmentation model that has the potential to save radiation oncologists a considerable amount of time. This could allow for the treatment of more patients, resulting in improved clinical practice efficiency and more reproducible contours.
\end{abstract}


\section{Introduction}
Radiotherapy (RT) plays a vital role in the treatment of various cancers, and accurate delineation of organs at risk (OARs) and target areas is essential for effective treatment planning~\cite{Brouwer2015}.
The delineation process involves defining the boundaries of healthy tissues (OARs) to be spared from high doses of radiation and tumor regions (targets) requiring irradiation.
Precise delineation ensures optimal radiation dose delivery to the tumor while minimizing damage to surrounding healthy tissues, leading to better treatment outcomes and reduced side effects for patients undergoing RT.

Acute Lymphocytic Leukemia (ALL) and Acute Myelogenous Leukemia (AML) are often treated with RT that targets malignant cells within the bone marrow, lymph nodes, and circulating blood.
Total Body Irradiation (TBI) is a widely used RT technique; however, it raises concerns due to radiation toxicity and late effects~\cite{mancosu2020}. Alternative treatments, such as Intensity Modulated Radiotherapy (IMRT), offer better control over radiation amount and location.
Recent advancements in the field have led to the development of specialized IMRT techniques, such as Total Marrow Irradiation (TMI) and Total Marrow and Lymph node Irradiation (TMLI)~\cite{mancosu2020}.
Compared to TBI, these treatments provide more precise radiation delivery, focusing on target volumes while minimizing exposure to healthy tissues~\cite{Wong2006}.
However, planning these treatments necessitates manual segmentation of several structures by ROs, including the Organs At Risk (OARs) and the Clinical Target Volume (CTV), which encompasses bones, marrow, spleen, and lymph nodes. The Planning Target Volume (PTV) is created by adding a margin to the CTV to account for uncertainties in setup positioning and internal motions~\cite{burnet2004}.

Manual contour quality is crucial for the success of IMRI techniques but poses a significant time burden on ROs, creating a need for automated segmentation of structures in Computerized Tomography (CT) scans.
Traditional auto-contouring methods, such as deformable image registration (DIR), probabilistic, and atlas-based methods, have contributed significantly to the field.
However, these methods often face limitations due to sensitivity to variations in patient anatomy, imaging artifacts, and reliance on predefined atlases that may not accurately represent individual patients.
Recent advances in deep learning for medical imaging segmentation have demonstrated potential for high-quality segmentation to support clinicians' work~\cite{wong2021implementation, harrison2022machine}.
While auto-contouring solutions exist and are integrated into commercial software like Raystation~\cite{raystationSegmentation} and Limbus AI~\cite{limbus}, they primarily focus on OAR segmentation and local PTV for specific tumor types.
Currently, no commercial software can segment PTV for TMLI treatment, and few studies in the literature have addressed this task.
In this paper, we present a deep learning-based auto-contouring method for segmenting PTV for TMLI treatment.

Specifically, we trained and compared two segmentation models based on the U-Net architecture~\cite{unet} with two different loss functions.
The models were trained and evaluated on a dataset of 100 patients treated with TMLI at the Humanitas Research Hospital between 2011 and 2021.
Our results demonstrate that, despite challenges in lymph node areas, the best model achieved a Dice score of $0.816 \pm 0.064$ for PTV segmentation.
These promising results represent a step towards developing a segmentation model for PTV that could potentially be effectively utilized in clinical practice, saving ROs significant time and enabling the treatment of more patients.

The paper is organized as follows.
In Section~\ref{sec:related}, we review the related work on PTV segmentation for TMLI treatment.
In Section~\ref{sec:methods}, we describe the dataset, the models, and the training and evaluation procedures.
Finally, in Section~\ref{sec:results}, we present the results of our experiments, and in Section~\ref{sec:conclusions}, we provide conclusions and outline future work directions.
\label{sec:introduction}

\section{Related Work}

Over the past decades, substantial efforts have been dedicated to developing reliable auto-contouring methods as an alternative to manual delineation. Early works on auto-contouring emphasized the use of deformable image registration (DIR), probabilistic, and atlas-based methods~\cite{langerak2013multiatlas, fritscher2014automatic, iglesias2015multiatlas, sharp2014vision, brock2017use, qazi2011autosegmentation, rigaud2019deformable}. In recent years, deep learning (DL) has emerged as a highly promising approach for creating auto-contouring tools effectively utilized in RT planning~\cite{wong2021implementation, harrison2022machine}. Numerous studies have successfully applied deep learning to auto-contouring of organs at risk (OARs) and target areas in various anatomical regions, such as the head and neck~\cite{Liu2020, Cardenas2020}, thorax~\cite{Yang2020, Feng2019}, abdomen~\cite{Tong2020, Kim2020}, and pelvis~\cite{Ma2022, Kalantar2021, Dong2019}.

Total marrow irradiation (TMI) poses a significant challenge as it requires delineation of OARs and targets across the entire patient's body. Regarding the segmentation of OARs, a few recent studies have demonstrated the feasibility of developing robust solutions for auto-contouring OARs in whole-body CT images. A notable example is the work of Chen et al.~\cite{CHEN2021175}, who developed WBNet, a system comprising multiple segmentation models~\cite{unet, Tang2019, 10.1007/978-3-319-46723-8_49} to segment 50 OARs across whole-body CT images. Zhou~\cite{Zhou2020} compared 2D and 3D convolution neural network (CNN) approaches in segmenting 17 structures in CT images, finding that the 3D approach achieved an average Dice Score (DSC) of 0.79, while the 2D approach reached 0.67. However, this improvement came at the cost of increased training and inference time. The framework used for 3D segmentation was divided into two steps: one that detected the volume of interest (VOI), a 3D bounding box containing the structure, and another that segmented the structure by focusing on the bounding box. Wasserthal et al.~\cite{Wasserthal2022} introduced a new model based on nnU-Net~\cite{nnUnet} to segment 104 anatomical structures (OARs, bones, muscles, and vessels) in whole-body CT images, achieving a DSC higher than 90\% for most structures. Additionally, commercial solutions such as the one developed by RayStation \cite{raystationSegmentation} and by Limbus AI \cite{limbus} are now available for segmenting OARs in CT images.

On the other hand, there are only a few authors that proposed approaches to segment the target of TMI.
Shi et al.~\cite{shi2022} proposed a dual encoder network that combines the Swin Transformer\cite{SwinTransformerLiu2021} and a ResNet~\cite{He2015}, and then uses a decoder to obtain the segmentation of the lymph nodes and bones in the CT image, which is an integral part of the Planning Target Volume (PTV). This new architecture achieved slightly better performance than the U-Net, U-Net++, and DeepLabV3+ in the segmentation tasks.
Additionally, Watkins et al.~\cite{Watkins2022} trained a commercial system based on U-Net to segment the PTV for TMI. Their model was trained to segment PTV-Bone, PTV-Lymph Nodes, PTV-ribs, and PTV-skull, achieving DSC of 0.851, 0.830, 0.946, 0.814, respectively.

\label{sec:related}

\section{Data}
The dataset employed in this study consists of images from 100 patients diagnosed with pathologically proven hematological malignancies. These patients were identified as candidates for allogeneic transplantation and received nonmyeloablative conditioning TMLI at Humanitas Research Hospital (Rozzano, Milan, Italy) between 2011 and 2021. All TMLI patients were immobilized in the supine position with their arms along their bodies, utilizing an in-house dedicated frame with multiple personalized masks~\cite{MANCOSU2021e98}. A free-breathing, noncontrast computed tomography (CT) scan with a 5-mm slice thickness was acquired for each patient using a BigBore CT system (Philips Healthcare, Best, Netherlands).

The TMLI clinical target volume (CTV) encompassed the bone marrow (CTV\_BM), spleen (CTV\_Spleen), and all lymph node chains (CTV\_LN). The CTV\_BM was considered equivalent to the skeletal bones, with the chest wall added to the ribs to account for breathing motions. To minimize oral cavity toxicity, the mandible was excluded from the CTV\_BM, along with the hands, which have an extremely limited bone marrow presence.
The total planning target volume (PTV\_Tot) was defined as the union of three PTVs, derived from the isotropic expansion of three corresponding CTVs, as follows: (i) PTV\_BM = CTV\_BM + 2 mm (+8 mm for arms and legs) to account for setup margin; (ii) PTV\_Spleen = CTV\_Spleen + 5 mm to account for breathing motions and setup margin; and (iii) PTV\_LN = CTV\_LN + 5 mm to account for target residual motion and setup margin.
\begin{figure}
    \begin{center}   
    \begin{tabular}{cc}
    \includegraphics[width=0.33\columnwidth]{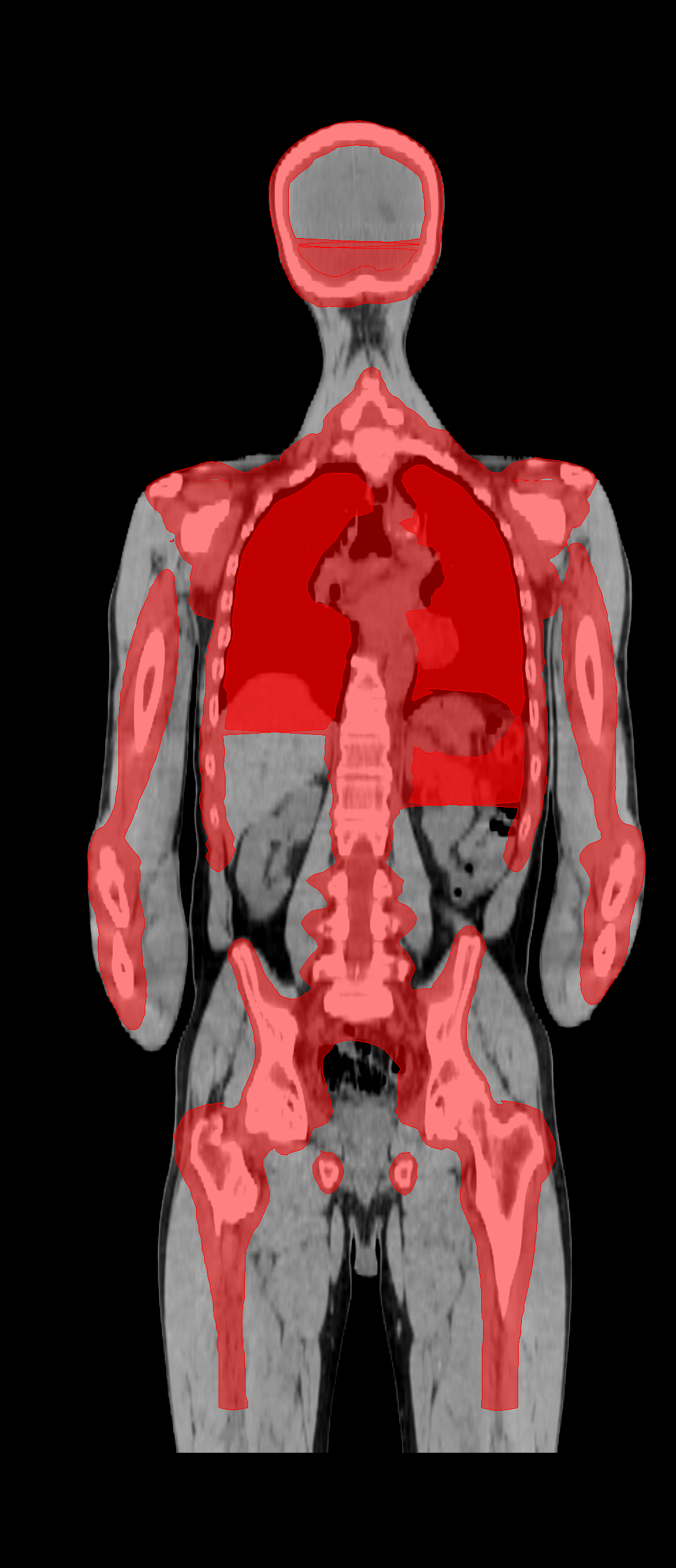} &
    \includegraphics[width=0.335\columnwidth]{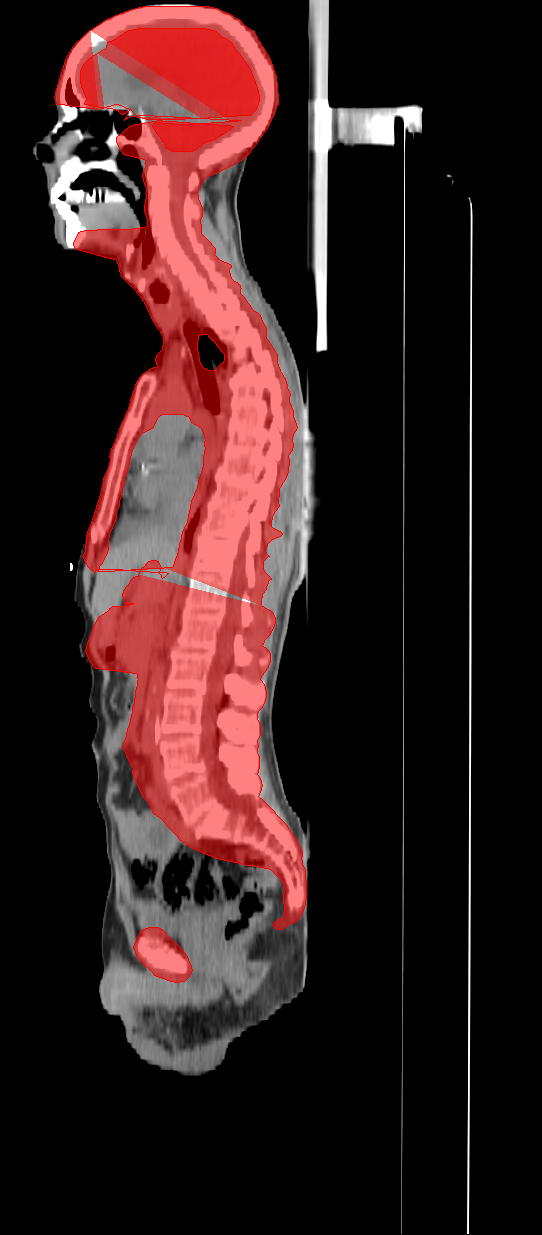}\\
    (a) & (b)
    \end{tabular}
    \end{center}
    \caption{An example of the PTV\_Tot: (a) coronal view and (b) saggittal view of a patient.}
    \label{fig:ptv}
\end{figure}    
Figure~\ref{fig:ptv} presents an example of the PTV\_Tot defined for a patient. 
The dataset contains the CT scans of each patient and the corresponding PTV\_Tot structure stored using the DICOM-RT format.
In this study, we trained a model to directly segment the PTV\_Tot target. Unlike other PTVs and CTVs, the PTV\_Tot is the only target consistently annotated by ROs for all 100 patients included in the dataset.


\label{sec:data}

\section{Methods and Experimental Design}
\label{sec:methods}
In this study, we trained a convolutional neural network (CNN) to segment the PTV\_Tot target from whole-body CT scans of patients. This section provides details on data preparation, the architecture of the convolutional network, and the loss function employed.

\subsection{Data Preparation}
We extracted from the DICOM-RT data of each patient the axial slices and their corresponding segmentation masks of PTV\_Tot. 
The number of axial slices per patient varies within a range from 164 to 534 and the slices are sized at 512 x 512 pixels with a resolution of 1.2 mm x 1.2 mm. 
The pixel values of the slices, which represent tissue radiointensity, are expressed in Hounsfield Units (HU). 
To enhance the contrast of all anatomical structures encompassed within the PTV, a linear Look Up Table (LUT) was applied to each pixel value, ranging from -160 HU to 240 HU, as suggested by the radiologist. 
Consequently, each slice was encoded as an 8-bit image, with pixel values ranging from 0 to 255.

\subsection{CNN Architecture}
The segmentation model used in this work is a U-Net\cite{unet}.
 The U-Net architecture consists of a contracting path and an expansive path. 
The network receives as input a 512 x 512 image of a CT slice and outputs a 512 x 512 segmentation mask.
It consists of the repeated application of two 3x3 convolutions (unpadded convolutions), each followed by a rectified linear unit (ReLU) and a 2x2 max pooling operation for downsampling. The expansive path consists of an upsampling of the feature map followed by a 2x2 deconvolution (“up-convolution”) that doubles the number of feature channels, a concatenation with the correspondingly cropped feature map from the contracting path, and two 3x3 convolutions, each followed by a ReLU.
The final layer is a 1x1 convolution that maps each 64-component feature vector to the desired number of classes. In total the network has 23 convolutional layers and slightly less than 8 million parameters.
%


\subsection{Loss Function}
In this work, we trained two models with two different loss functions, the Binary Cross Entropy Loss (BCEL) \cite{burnet2004} and the Dice Loss (DL) \cite{sudre2017}, as described in Equation \ref{Eq:BCEL} and Equation \ref{Eq:DL}, respectively. In both equations, the $p$ represents each pixel of the prediction and the $y$ represents the pixels of the groundtruth.
\begin{equation}
    BCE(p,y) = -(y \log(p) + (1 - y) \log(1 - p)) 
    \label{Eq:BCEL}
\end{equation}

\begin{equation}
    DL(p,y) = 1 - \frac{2yp + 1}{y + p + 1}
    \label{Eq:DL}
\end{equation}

BCEL calculates the loss for individual pixels, while DL computes the overlap between the ground truth and prediction, addressing class imbalance by considering only the percentage of prediction overlapping with the ground truth rather than the size of the prediction.

\subsection{Training Process}
\begin{figure}[hbt!]
     \centering
         \includegraphics[width=0.45\textwidth]{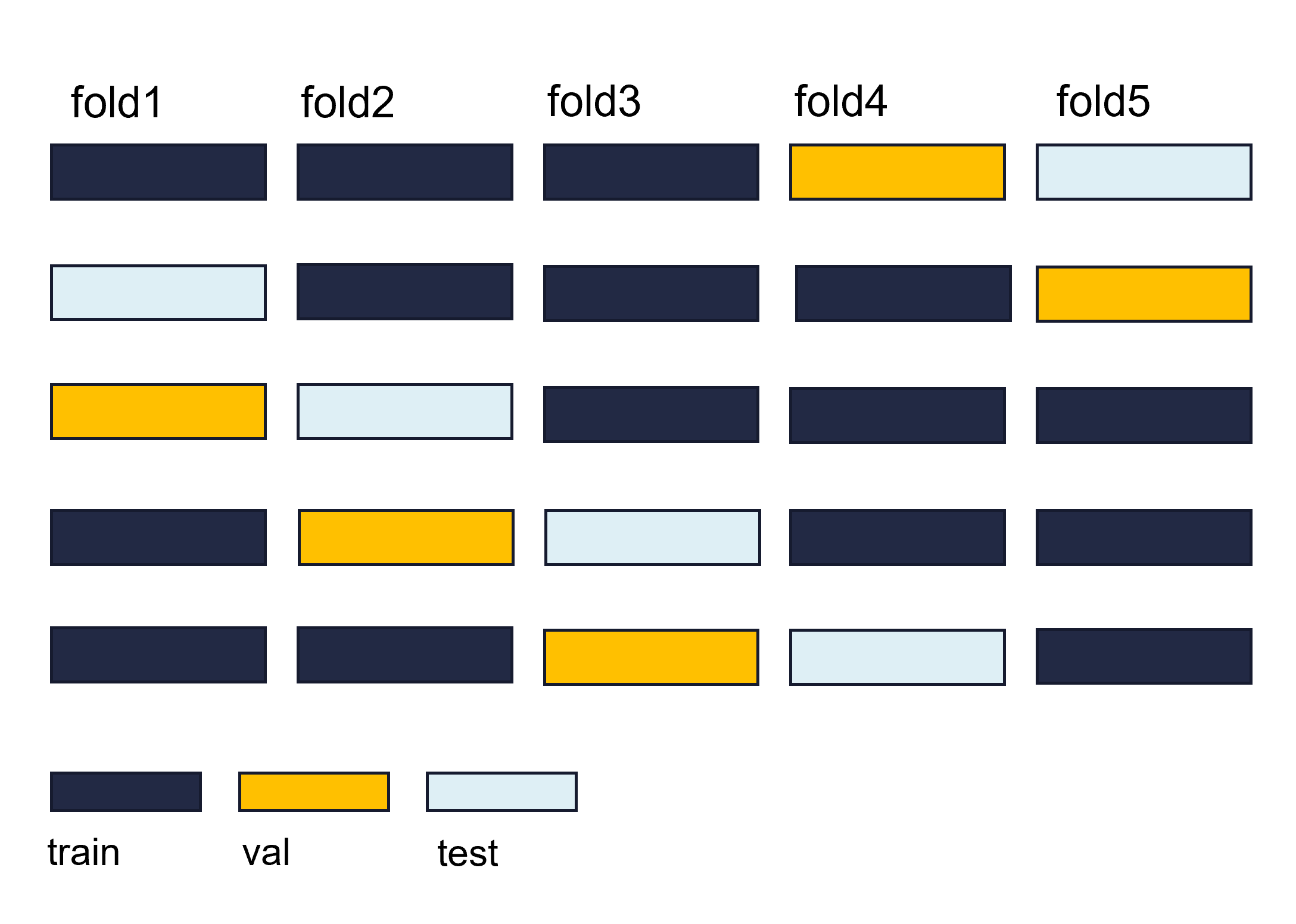}
         \caption{The cross-validation scheme used to assess the performance of the trained models.}
         \label{fig:crossValidation}
\end{figure}

To train and evaluate the performance of the segmentation models, we employed a 5-fold cross-validation method as depicted in Figure~\ref{fig:crossValidation}. 
The dataset is divided into five equally-sized folds (20 patients per fold), with three folds allocated for training, one for validation, and one for testing. 
To account for variations in the delineation process and equipment, we ensured that the five folds were uniformly sampled concerning the acquisition dates of the CT series. 
Training was conducted with a batch size of 4 and a learning rate of $10^{-5}$. 
The training process was terminated as soon as no significant changes are observed in the validation loss for 10 consecutive epochs.

\subsection{Evaluation Metrics}
To assess the segmentation performance of the models, we calculated the Dice Score (DSC) and Hausdorff Distance (HD) as described in Equation \ref{Eq:DSC} and Equation \ref{Eq:HD}, respectively. 
The DSC measures the overlap between the ground truth and the predicted mask, with $X$ representing the set of positive pixels in the ground truth and $Y$ representing the set of positive pixels in the prediction.

\begin{equation}
DSC = \dfrac{2 |X \cap Y| }{|X| + |Y|}
\label{Eq:DSC}
\end{equation}

The HD measures the maximum distance of all the nearest distances between the surfaces of the two sets $X$ and $Y$, denoted as $S_X$ and $S_Y$. To mitigate the impact of outliers on HD values, we employed HD95, which excludes the top 5\% highest HD values.

\begin{equation}
	HD=max\left\{\max_{x\in S_X} d(x,S_Y),
    \max_{y\in S_Y} d(y,S_X) \right\},
    \label{Eq:HD}
\end{equation}




\section{Results}
In this section, we present and discuss the results of two models trained to segment the Planning Target Volume (PTV\_Tot) using two different loss functions, Dice Loss (DL) and Binary Cross-Entropy Loss (BCEL). We evaluate their effectiveness in the segmentation process and provide visual examples of segmentation outcomes to identify and understand the areas where the model struggled to perform accurate segmentation.

\begin{figure}
    \centering 
    \begin{subfigure}{0.45\columnwidth}
      \includegraphics[width=\linewidth]{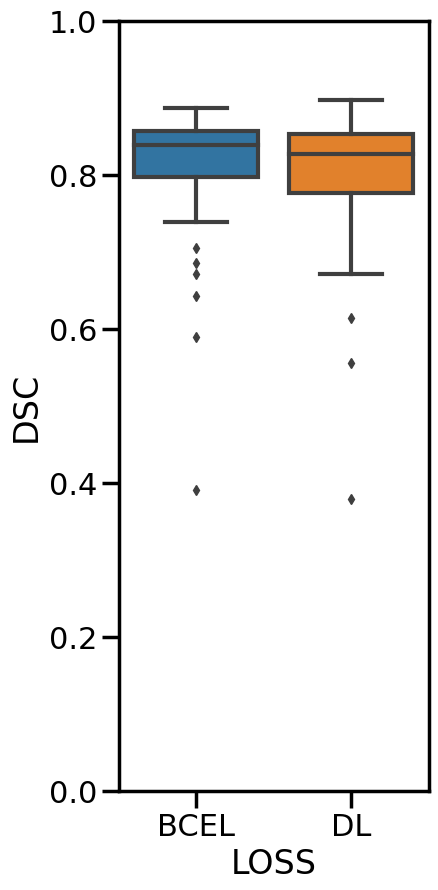}
      \label{Fig:dsc}
    \end{subfigure}\hfil 
    \begin{subfigure}{0.45\columnwidth}
      \includegraphics[width=\linewidth]{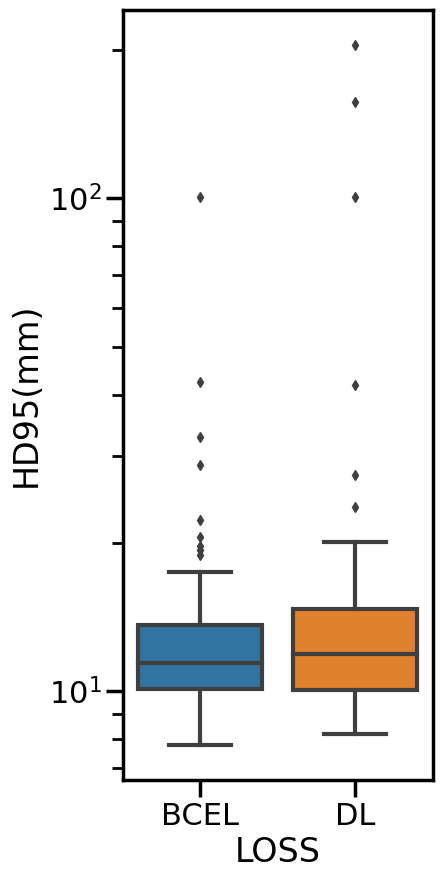}
      \label{Fig:hd95}
    \end{subfigure} 
\caption{Boxplot of the DSC (left) and the HD95 values (right) of the BCEL and DL models.}
\label{Fig:loss_comparison_dsc_hd95}
\end{figure}

Figure \ref{Fig:loss_comparison_dsc_hd95} compares the segmentation performance of the models trained with BCEL and DL loss functions. 
The box plot on the left side presents the values of the Dice Similarity Coefficient (DSC), and the box plot on the right presents the 95\% Hausdorff Distance (HD95) values, along with their standard deviations, for all patients across all folds. 
The BCEL-trained model consistently outperforms the DL-trained model in the two metrics. Notably, the DSC values for the BCEL model (0.816 $\pm$ 0.064) are higher than those for the DL model (0.806 $\pm$ 0.072), indicating a better overlap between the predicted segmentation and the ground truth. Additionally, the BCEL model exhibits a lower average HD95 value (13.81 mm $\pm$ 7.602 mm) compared to the DL model (17.575 mm $\pm$ 17.386 mm).
The range of the patient's DSC values in the BCEL model is from 39\% to 89\%, and in the DL model is from 38\% to 90\%, although, in the HD95 values the BCEL model range is from 7mm to 100 mm and in the DL model is from 8 mm to 205 mm.
Lower HD95 values suggest that the BCEL model provides a more accurate segmentation in terms of the maximum and 95\% percentile distances between the prediction and the ground truth.
It is also possible to notice that the performances of the BCEL are more consistent, with a lower spread and fewer outliers.



\begin{figure*}

    \centering 
    \begin{subfigure}{0.23\textwidth}
      \includegraphics[width=\linewidth]{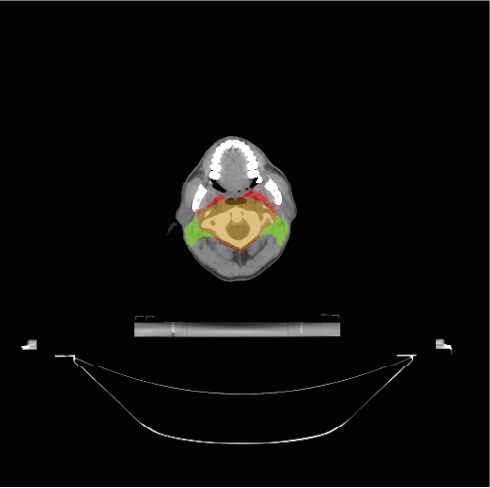}
      \label{Fig:loss_comparison_worst_a}
    \caption{}
    \end{subfigure}\hfil 
    \begin{subfigure}{0.23\textwidth}
      \includegraphics[width=\linewidth]{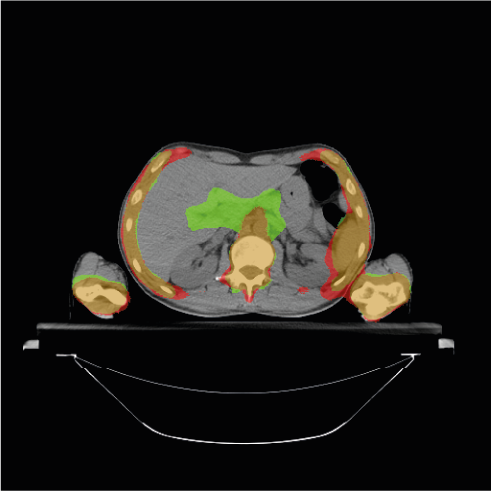}
      \label{Fig:loss_comparison_worst_b}
      \caption{}
    \end{subfigure} 
    \begin{subfigure}{0.23\textwidth}
      \includegraphics[width=\linewidth]{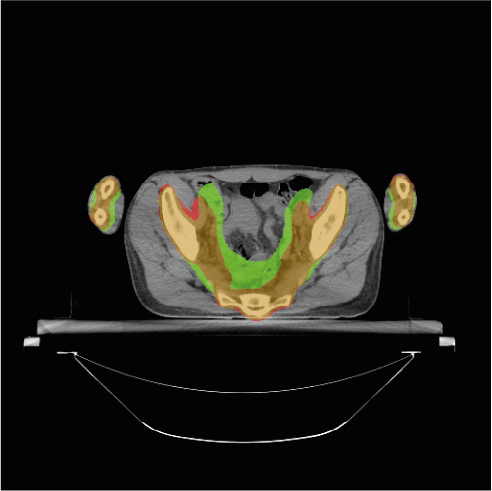}
      \label{Fig:loss_comparison_worst_c}
      \caption{}
    \end{subfigure}\hfil
    \begin{subfigure}{0.23\textwidth}
      \includegraphics[width=\linewidth]{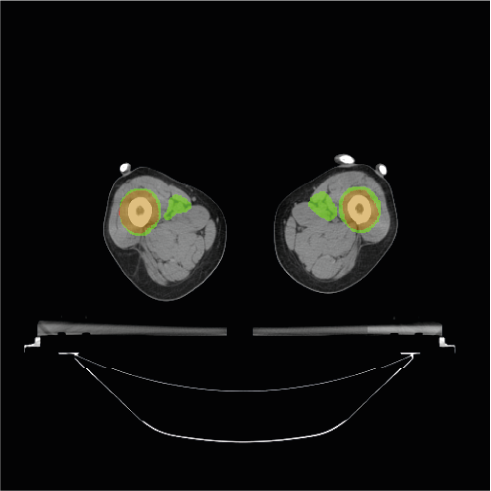}
      \label{Fig:loss_comparison_worst_d}
      \caption{}
    \end{subfigure}
    \medskip
    
    \begin{subfigure}{0.23\textwidth}
      \includegraphics[width=\linewidth]{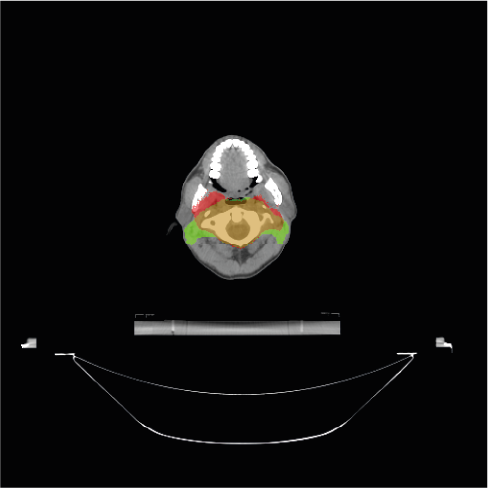}
      \label{Fig:loss_comparison_worst_e}
    \caption{}
    \end{subfigure}\hfil 
    \begin{subfigure}{0.23\textwidth}
      \includegraphics[width=\linewidth]{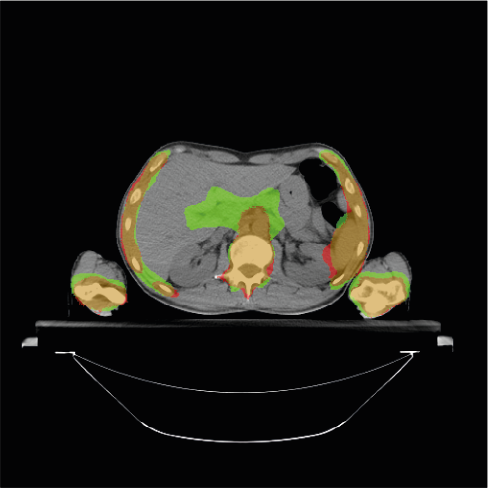}
      \label{Fig:loss_comparison_worst_f}
      \caption{}
    \end{subfigure} 
    \begin{subfigure}{0.23\textwidth}
      \includegraphics[width=\linewidth]{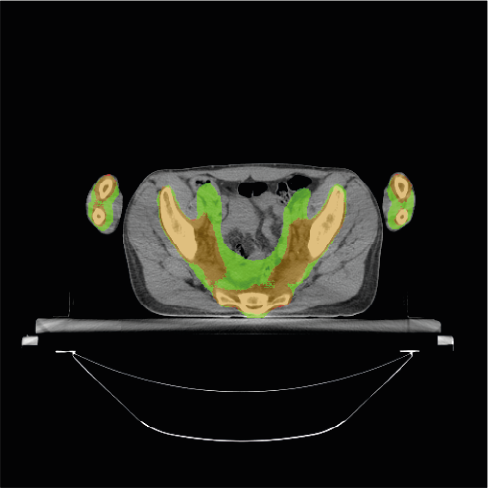}
      \label{Fig:loss_comparison_worst_g}
      \caption{}
    \end{subfigure}\hfil
    \begin{subfigure}{0.23\textwidth}
      \includegraphics[width=\linewidth]{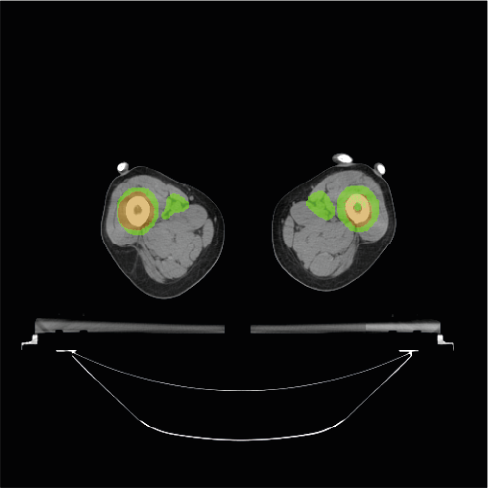}
      \label{Fig:loss_comparison_worst_h}
      \caption{}
    \end{subfigure}
\caption{Examples of poorly segmented CT slices. In each slice, the ground truth of the PTV is overlayed in green, while the prediction is overlayed in red.
The first row shows the slices with the predictions of the BCEL model, while the second row the slices with the predictions of the DL model. 
Accordingly, in each slice, false negative regions appear in green, false positive regions appear in red, and true positive regions appear in yellow.}
\label{Fig:loss_comparison_worst}
\end{figure*}
To further understand the areas where the models do not accurately segment the PTV, we provide some visual examples of the poor segmentations Figure~\ref{Fig:loss_comparison_worst}.
The first row of the Figure~\ref{Fig:loss_comparison_worst} shows four examples of slices poorly segmented by the BCEL model.
The first slice (Figure~\ref{Fig:loss_comparison_worst}a) corresponds to the head and neck area. The segmentation error in the left and right areas of the head (i.e., the green areas) may be due to conservative segmentation of the second level of neck lymph nodes in the ground truth. The standard reference point for segmenting these lymph nodes might be a slightly higher slice, but the radiation oncologist likely included the target on this slice due to the large CT thickness of 5 mm.
The second slice (Figure~\ref{Fig:loss_comparison_worst}b) shows an absence of hepatic portal lymph nodes. This area is subject to high anatomical variability, and due to the lack of contouring guidelines~\cite{Dei2023} in the early years of data acquisition, this area was often omitted by the radiation oncologist.
In the third slice (Figure~\ref{Fig:loss_comparison_worst}c), the absence of the pre-sacral lymph nodes (anterior to the sacrum) is evident. Moreover, the predicted segmentation of the external iliac lymph nodes is inaccurate. This target area is defined by adding a margin to the iliac vessels.
In the fourth slice (Figure~\ref{Fig:loss_comparison_worst}d), representing the legs area, the prediction fails to segment the inguinal lymph nodes, likely because these are frequently omitted by physicians.
The second row of the Figure~\ref{Fig:loss_comparison_worst} shows the same slices overlayed with the predictions of the DL model. 
Notably, the segmentation errors of the DL model closely resemble those of the BCEL one, previously discussed.
Please notice that the segmentation mistakes of DL model are very similar to the ones of the BCEL model, previously discussed. 
Nevertheless, a higher propensity for false negatives is discernible in the DL model's segmentations, evidenced by the larger green regions surrounding the bones in Figure~\ref{Fig:loss_comparison_worst}h and within the iliac lymph node area in Figure~\ref{Fig:loss_comparison_worst}g.


\label{sec:results}

\section{Conclusions}
In this study, we trained two U-Net segmentation models with different loss functions to segment the Planning Target Volume (PTV) for Total Marrow Irradiation (TMLI) treatment. The models were trained on a dataset comprising 100 patients treated at Humanitas Research Hospital from 2011 to 2021. 
The PTV encompasses several complex structures that are subject to anatomical variability and includes an extra margin according to specific clinical guidelines. 
This increased complexity poses challenges for accurate PTV segmentation, with the most notable difficulties arising in the lymph node areas. 
Moreover, the visual inspection of segmentation errors indicates that the considerable variability in the ground truth, stemming from the absence of comprehensive delineation guidelines, may have influenced the performance of the models.
Despite these challenges, the performance of the best model is promising, with Dice Similarity Coefficient (DSC) values of $0.816 \pm 0.064$. 

We believe that our results are encouraging and represent a step towards the development of a segmentation model for PTV that could potentially be effectively used in clinical practice, saving a significant amount of time and leading to more robust contours.
Future research will involve exploring additional deep learning architecture and enhancing the model input by incorporating the delineation of pertinent anatomical structures for PTV segmentation. 
Furthermore, we intend to conduct an in-depth examination of the groundtruth in the dataset employed in this study to improve adherence to delineation guidelines.

\label{sec:conclusions}


\bibliographystyle{IEEEtran} 
\bibliography{main.bib}

\end{document}